\documentclass{article} 
\usepackage{iclr2026_conference,times}

\usepackage{amsmath,amsfonts,bm}

\def\eqref#1{equation~\ref{#1}}

\def\1{\bm{1}}

\DeclareMathAlphabet{\mathsfit}{\encodingdefault}{\sfdefault}{m}{sl}
\SetMathAlphabet{\mathsfit}{bold}{\encodingdefault}{\sfdefault}{bx}{n}

\usepackage[pdftex]{graphicx}
\usepackage{hyperref}
\usepackage{url}
\usepackage{multirow}
\usepackage{listings}

\newcommand{\ccol}[2]{ \multicolumn{#1}{c}{#2}}

\title{the Mouth is Not the Brain: Bridging Energy-Based World Models and Language Generation}

\author{Junichiro Niimi \\
Faculty of Business Management\\
Meijo University\\
Nagoya, Aichi 4688502, Japan \\
\texttt{jniimi@meijo-u.ac.jp} 
}

\iclrfinalcopy 
\begin{document}

\maketitle

\begin{abstract}
Large Language Models (LLMs) generate fluent text, yet whether they truly understand the world or merely produce plausible texts about it remains contested. We propose an architectural principle, the mouth is not the brain, that explicitly separates world models from language models. Our architecture comprises three components: a DBM that captures domain structure as an energy-based world model, an adapter that projects latent belief states into embedding space, and a frozen GPT-2 that provides linguistic competence without domain knowledge. We instantiate this framework in the consumer review domain using Amazon smartphone reviews. Experiments demonstrate that (1) world model conditioning achieves lower cross-entropy loss and higher semantic similarity than architectural baselines including direct projection and full fine-tuning, while qualitative analysis reveals that soft prompt conditioning resolves a trade-off that prompt-based approaches cannot: simple prompts lack expressiveness while detailed prompts cause output collapse in small LLMs; (2) the DBM's energy function distinguishes coherent from incoherent market configurations, assigning higher energy to implausible brand-price combinations; and (3) interventions on specific attributes propagate causally to generated text with intervened outputs exhibiting distributions statistically consistent with naturally occurring samples sharing the target configuration. These findings suggest that even small-scale language models can achieve consistent, controllable generation when connected to an appropriate world model, providing empirical support for separating linguistic competence from world understanding.
\end{abstract}

\section{Introduction}
Large Language Models (LLMs) have emerged as remarkably capable language generators, with sophisticated reasoning techniques such as chain-of-thought prompting \citep[CoT;][]{cot} and multi-agent deliberation \citep{llm_debate} extending far beyond simple next-token prediction. However, a fundamental question remains: do these models understand the world, or do they merely generate plausible texts about it? The underlying representations emerge from language: they are trained on language and optimized for language.

This concern has gained attention, with LeCun describing autoregressive LLMs as ``doomed'' due to their inability to model causality\footnote{Joint Mathematics Meetings, January 2025.}, and Hassabis emphasizing that artificial general intelligence (AGI) requires world models capable of understanding physical reality\footnote{Google DeepMind Podcast, December 2025.}. LLMs learn to talk about the world, but whether they have formed a coherent model of it is far from clear. The mouth speaks fluently; whether the brain comprehends is another matter. This motivates our architectural principle: \textit{the mouth is not the brain}.

This paper therefore proposes to separate the world model from the language model. The world model learns the latent structure of a domain (e.g., what entities exist, how they relate, and which configurations are coherent) while the language model renders this structured representation as fluent text. We instantiate this principle in consumer product reviews, where structured behavioral data coexists with rich textual expressions.

To rigorously evaluate this architecture, we use deliberately minimal components: a Deep Boltzmann Machine \citep[DBM;][]{dbm} as the world model and a frozen GPT-2 as the language model. This choice is intentional. Frontier LLMs may have already internalized implicit world models, making it impossible to disentangle the contribution of explicit world modeling from the LLM's own capabilities. By pairing a language model with limited capacity against an explicit energy-based world model, we can cleanly isolate the causal role of world model conditioning, ensuring that performance gains stem from structured understanding rather than the language model's latent knowledge. This constitutes an ablation study of the separation principle itself.

\section{Related Work}
\subsection{Language Model}
Large-scale autoregressive models such as GPT have achieved remarkable fluency across diverse tasks, and recent advances, such as CoT, multi-agent frameworks, and reinforcement learning from human feedback \citep[RLHF;][]{rlhf,reinforce1}, have extended their capabilities further. However, purely text-based learning faces fundamental hurdles. LeCun has argued that scaling autoregressive models alone cannot yield human-level intelligence \citep{lecun2022}, and similar concerns have been raised about the gap between fluent generation and genuine understanding \citep{bender_acl2020}. Studies on hallucination further support this critique, identifying it as a structural limitation of LLMs \citep[e.g.,][]{hallucination,hallucination_inevitable1,niimi2025hallucinate}. The reliance on next-token prediction provides no direct incentive to model causal structure or support counterfactual reasoning. While emergent world representations have been observed within sequence models \citep{othellogpt}, such representations remain implicit and entangled with language generation; our work instead pursues explicit architectural separation.

These limitations become concrete in application domains such as consumer review generation, where prior work constructs prompts from user behavioral history \citep{generate_review_llm1} yet struggles with latent dimensions of consumer heterogeneity (e.g., price sensitivity, brand loyalty, lifestyle preferences) that are difficult to articulate explicitly \citep{personalLLM,llm_persona}. Such difficulties motivate the exploration of complementary approaches: world models that represent domain structure independently of language.

\subsection{World Model}
World models have gained attention as a complementary approach to LLMs, with two major trends: RL-based approaches that learn environmental dynamics through state prediction \citep{schmidhuber2015,ha2018}, and representation-learning approaches such as JEPA that capture semantic structure in abstract embedding spaces \citep{lecun2022,i-jepa,v-jepa}. Both traditions share a key principle: separating world understanding from downstream output generation, routing information through an abstract bottleneck representation. LeCun has positioned Energy-based Models (EBM) as a natural framework for this purpose, where an energy function distinguishes plausible from implausible configurations \citep{lecun2022}.

Our approach adopts this energy-based formulation, but differs in a key respect: rather than modeling temporal dynamics, we capture \textbf{co-occurrence structure} (which attribute configurations are coherent). This design reflects our application domain, where ``world structure'' is better characterized by market regularities than by sequential state evolution.

\subsection{Deep Boltzmann Machine}
We employ DBMs as our world model component. DBMs are EBMs where the probability of a configuration is determined by an energy function:
\begin{align}
E(v, h) = -v^\top W h - b^\top v - c^\top h
\end{align}
where $v$ represents visible units, $h$ hidden units, and lower energy indicates higher probability: $P(v) \propto \sum_h \exp(-E(v, h))$. DBMs stack multiple hidden layers with undirected connections, enabling hierarchical structure learning through both bottom-up and top-down inference.

A key property for our purposes is the unsupervised training objective: the model learns statistical regularities without task-specific supervision, constructing a representation of ``how the world is.'' We use the term ``belief'' in two senses: the technical sense (posterior over hidden units) and the cognitive sense (propositional attitudes). While DBMs have been applied to multimodal learning \citep{multimodal,multimodal_dbm}, prior work does not address connection to external language models. Our contribution lies in using the DBM as an explicit, inspectable world model that conditions a frozen LLM.

\subsection{Combination of LLMs and World Models}
Recent work has explored LLM-world model combinations, including social simulation systems \citep{simulacra,opencity} and JEPA-style auxiliary losses \citep{llm-jepa}, but in these systems the world model remains implicit.  In contrast, our architecture employs an explicit, separately trained world model that conditions a frozen LLM through learned adapters, maintaining clear separation between domain understanding and language generation.

\section{Proposed Model}
\begin{figure}[htb]
\begin{center}
   \includegraphics[width=0.8\linewidth]{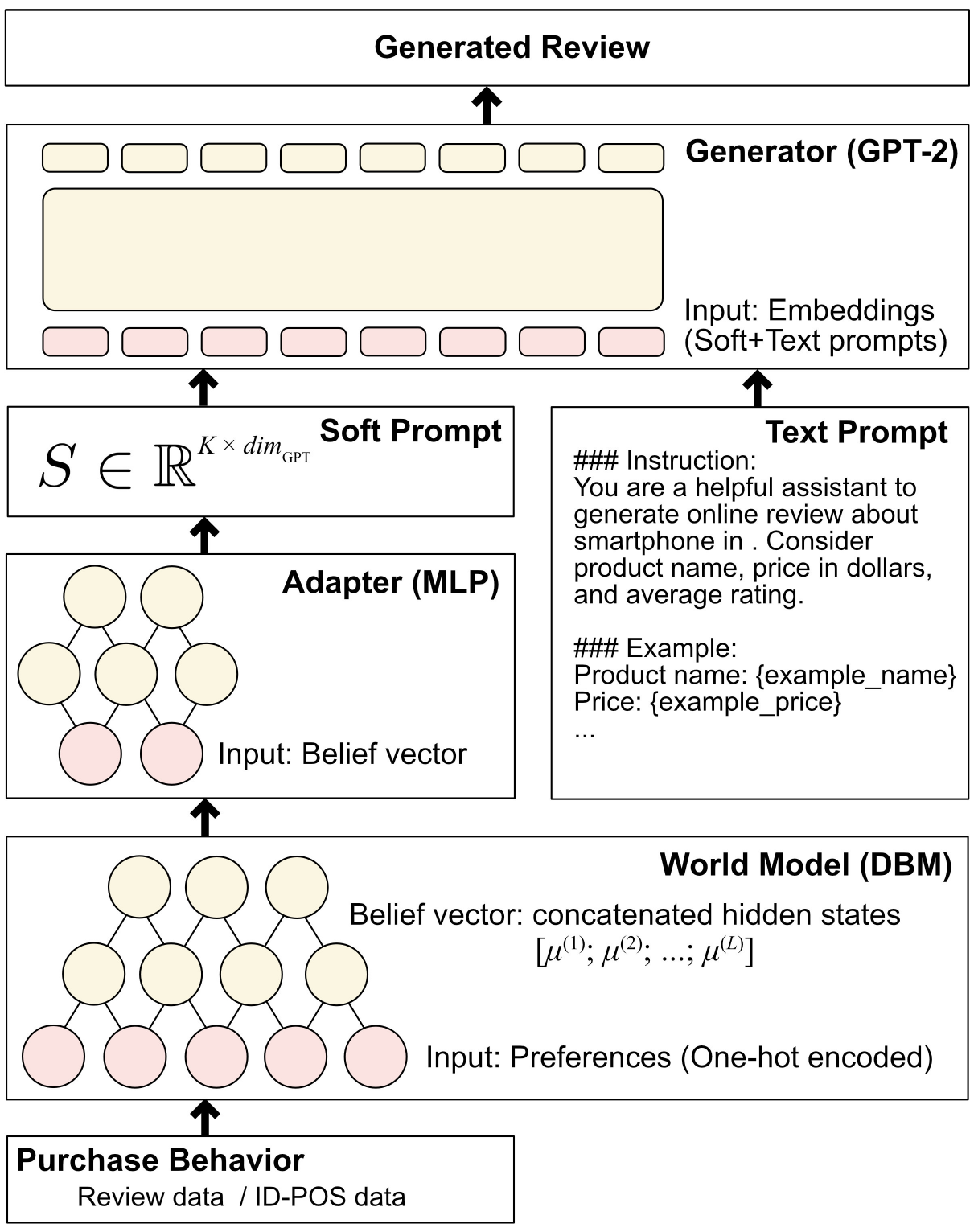}
   \caption{Model architecture}\label{fig:architecture}
\end{center}
\end{figure}

Our proposed model consists of three components (Fig.~\ref{fig:architecture}): a world model that captures domain structure, an adapter that projects latent representations into language space, and a language model that generates text. We first conduct DBM's layer-wise pretraining and joint fine-tuning, followed by training of the adapter. The language model is frozen through training and inference.

\subsection{World Model: Deep Boltzmann Machine}
The world model is implemented as a DBM with one visible layer ($J$ units) and $L=2$ hidden layers ($H_l$ units each). The visible layer represents consumer behavior as a one-hot encoded binary vector (brand, price tier, rating, contextual signals). The energy function is:
\begin{align}
E(v, h^{(1)}, \ldots, h^{(L)}) = -\sum_{l=1}^{L} h^{(l-1)\top} W^{(l)} h^{(l)} - b^\top v - \sum_{l=1}^{L} c^{(l)\top} h^{(l)}
\end{align}
where $h^{(0)} \equiv v$, and $W^{(l)}$, $b$, $c^{(l)}$ are weight matrices and biases. The bipartite structure enables efficient mean-field inference:
\begin{align}
\mu^{(l)} \leftarrow \sigma\left( W^{(l)\top} \mu^{(l-1)} + W^{(l+1)} \mu^{(l+1)} + c^{(l)} \right)
\end{align}
where $\mu^{(0)} \equiv v$ and $W^{(L+1)} \mu^{(L+1)} \equiv 0$.

Training follows the standard two-phase procedure \citep{dbm}: layer-wise pretraining with RBMs using Contrastive Divergence, followed by joint fine-tuning with Persistent Contrastive Divergence. The concatenated hidden activations $[\mu^{(1)}; \ldots; \mu^{(L)}]$ constitute the belief representation passed to the adapter.

\subsection{Language Model: Frozen GPT-2}
We employ GPT-2 as the language model, with all parameters frozen throughout training and inference. GPT-2 serves as a language faculty: it does not determine the content of generation but provides the grammatical and stylistic competence necessary to articulate beliefs as fluent text. Conditioning comes not from textual prompts but from soft prompt embeddings projected from the world model's latent state.

By keeping GPT-2 frozen, we enforce a strict separation: the world model learns domain structure, while the language model contributes only linguistic competence. The language model never observes raw tabular features. This ensures that content-level variation in generated text originates from the world model, not from the language model's internal representations (See Appendix \ref{sec:prompt} for the prompt).

\subsection{Adapter: Projecting Beliefs to Language Space}
The adapter bridges the world model and the language model by projecting latent belief states into GPT-2's embedding space. The concatenated mean-field activations $[\mu^{(1)}; \ldots; \mu^{(L)}]$ are passed through a multi-layer perceptron \citep[MLP;][]{mlp} that outputs $K$ soft prompt embeddings $S \in \mathbb{R}^{K \times D}$, prepended to text prompt embeddings as conditioning context.

This setup follows the soft prompt tuning paradigm \citep{soft_prompt}, but differs in a key respect: our soft prompts are derived from an external world model rather than learned end-to-end from text alone. During training, only the adapter parameters are updated to minimize cross-entropy loss between generated and ground-truth reviews with AdaMax \citep{adam}, while DBM weights are fixed after finetuning and GPT-2 remains frozen.

\subsection{Inference and Dataset}
We use Amazon product reviews \citep{amazon_product_review2,amazon_product_review1} for smartphones. English reviews of verified purchases are extracted using FastText \citep{fasttext1,fasttext2}, limited to 100--1024 characters ($n=55,000$; train/val/test = 52,952/1,024/1,024).

A consumer profile is represented as a binary visible vector $v \in \{0, 1\}^{160}$, encoding one-hot features (brand, price tier, rating) and binary contextual signals (repeat buyer, accessory purchases, etc.) derived from past reviews. Given $v$, we infer the latent belief state via mean-field iteration until convergence, then pass the concatenated activations $[\mu^{(1)}; \ldots; \mu^{(L)}]$ through the adapter to produce soft prompt embeddings. These are prepended to a text prompt containing a brief instruction and one-shot example, and fed to frozen GPT-2 (temperature=0.7, max\_new\_tokens=100) for autoregressive generation.

\section{Experiment I: Generation Quality}
\subsection{Method}
To verify text generation from both quantitative and qualitative perspectives, we conduct two analyses. First, for qualitative validation, we compare prompt-based approaches---Baseline 1 (GPT-2 with text prompt only) and Baseline 2 (GPT-2 with extended prompt including detailed behavioral context)---against the proposed model (DBM $\to$ adapter $\to$ GPT-2), generating outputs from the same input.

Second, for quantitative validation, we test whether conditioning through the world model produces text more consistent with ground-truth reviews (\textbf{H1}). We compare the proposed model against three architectural baselines on 500 test samples: B0 (frozen GPT-2 with one-shot ICL, no training), B1 (tabular $\to$ MLP adapter $\to$ frozen GPT-2, bypassing the DBM), and B2 (GPT-2 with all parameters fine-tuned on domain data). B1 isolates whether the DBM's latent representation is necessary, while B2 tests whether full fine-tuning can substitute for the separation architecture. We evaluate using cross-entropy (CE) loss and cosine similarity computed with SBERT (all-MiniLM-L6-v2).

\subsection{Results and Discussions}

The results of qualitative validation are shown in Table~\ref{tab:generated}. Baseline 1 reflected the positive rating but included minor complaints (speaker volume, screen size) despite the high rating and positive tags such as ``overall" and ``amazing". Baseline 2 mentions product quality and price, but the content remains generic and the output collapses due to prompt length exceeding GPT-2's effective context window. The proposed model successfully produces coherent content reflecting the input topics (battery and screen) with appropriate sentiment. Furthermore, it references the renewed/refurbished nature of the product, framing its recommendation around quality rather than price expectations. This demonstrates that compressing behavioral information through a world model enables controlled generation even with small-scale language models.

\begin{table}[htb]
\centering
   \caption{Qualitative evaluation: generated results (based on the same input)}\label{tab:generated}
   \scalebox{0.85}{
   \begin{tabular}{
@{}cl@{}
}
\hline
\textbf{Input Tags} & Price=mid, Brand=Apple, Rating=5, Topics=[battery, screen, renewed, overall, amazing] \\
\hline
\multirow{4}{*}{\shortstack{\textbf{Baseline 1}\\{\small (text prompt)}}}
&\\
& I am happy with the quality of this Apple iPhone 11. \\
&Although one minor complaint is the speaker is too loud and the screen is too small.\\
&\\
\hline
\multirow{4}{*}{\shortstack{\textbf{Baseline 2}\\{\small (text + belief)}}}
&Product description:\\
&You find a good product with excellent price.\\
&Review:\\
&Review:\\
\hline
\multirow{4}{*}{\shortstack{\textbf{Proposed Model}\\{\small (Boltzmann-GPT)}}}
&The phone is gorgeous, the screen is amazing, and everything is working normally. I don’t\\
&have any issues with battery life and it’s so far perfect. It’s definitely a great buy.  I wouldn’t \\
&recommend this phone if you’re looking for a refurbished phone, but if you’re looking for a\\
&great one then you’re going to love this one. Love the screen!\\
\hline
   \end{tabular}
   }
\end{table}

These results reveal a fundamental trade-off in prompt-based conditioning for small language models: simple prompts cannot convey complex information, while detailed prompts cause the model to collapse. The proposed architecture resolves this trade-off through soft prompt conditioning, enabling stable transmission of rich behavioral signals that would otherwise be either omitted or cause catastrophic failure\footnote{While Table \ref{tab:generated} demonstrates high-rating generation, the architecture also controls low-rating generation effectively (Appendix \ref{sec:appendix_negative_reviews}).}.

Table \ref{tab:exp1} shows quantitative results on the test set. The proposed model achieves the lowest CE loss (3.32) and highest cosine similarity (0.43), outperforming all baselines. B0 (frozen ICL) establishes the lower bound, confirming that domain adaptation through the world model yields substantial gains. B1 (direct MLP) improves over B0 in both metrics, but the DBM's latent representation provides further gains in CE loss (${\sim}0.2$) and cosine similarity (${\sim}0.04$), indicating that the structured belief representation contributes beyond what a direct projection can achieve. B2 (fine-tuned LLM) produces the worst CE loss (4.74) due to severe overfitting, yet its cosine similarity (0.40) exceeds B0 and B1, suggesting that fine-tuning captures some semantic structure despite loss degradation. The separation architecture (frozen LLM + trained adapter) remains a more effective inductive bias overall. These results support H1.

\begin{table}[htb]
\begin{center}
\caption{Quantitative evaluation on the test set ($n=500$). CE = cross-entropy; CosSim = cosine similarity (SBERT).}\label{tab:exp1}
\begin{tabular}{lrr}
\hline
\ccol{1}{\bf Model} & \ccol{1}{\bf CE Loss~} & \ccol{1}{\bf CosSim}\\
\hline
Proposed: DBM $\to$ Adapter $\to$ GPT-2 & \textbf{3.32} & \textbf{0.43} \\
B0: Frozen ICL (no training) & 3.59 & 0.38 \\
B1: Direct MLP (no DBM) & 3.52 & 0.39 \\
B2: Fine-tuned LLM & 4.74 & 0.40 \\
\hline
\end{tabular}
\end{center}
\end{table}

\section{Experiment II: World Model Coherence}
\subsection{Method}
Experiment I evaluated the integrated model but did not reveal whether the DBM itself has learned meaningful structure. We therefore test two hypotheses, \textbf{H2: Energy changes under intervention are consistent between training and held-out test samples} and \textbf{H3: The DBM assigns higher energy to configurations that violate learned market structure}. 

We compute variational free energy $\tilde{F}(v)$ for each configuration:
\begin{equation}
\tilde{F}(v) = -\sum_{l=1}^{L} \mu^{(l-1)\top} W^{(l)} \mu^{(l)} - b^\top v - \sum_{l=1}^{L} c^{(l)\top} \mu^{(l)} + \sum_{l=1}^{L} H(\mu^{(l)})
\end{equation}
where $H(\mu^{(l)})$ denotes entropy terms. We measure energy changes under brand--price interventions: samples from Mid, High, and Premium tiers are intervened to Entry tier, comparing Apple (strongly associated with premium pricing) against Others (minor brands with heterogeneous pricing). The percentage change $\Delta F = (\tilde{F}(v_{\text{intervened}}) - \tilde{F}(v_{\text{original}})) / \tilde{F}(v_{\text{original}}) \times 100$ indicates whether the DBM recognizes intervened configurations as implausible.

\subsection{Results and Discussions}
\subsubsection{Generalizability of DBM}
Table \ref{tab:dbm_generalizability} shows that energy changes under intervention are highly consistent between training and test sets. Both the rank ordering and absolute magnitudes are preserved across all conditions, indicating that the DBM has learned generalizable co-occurrence structure rather than memorizing training samples.

\begin{table}[htb]
\centering
   \caption{The mean difference of energy under the simple intervention for training and test sets.}\label{tab:dbm_generalizability}
   \begin{tabular}{@{}clccrr@{}}
\hline
&\ccol{3}{\bf Intervention}& \ccol{1}{\bf Train}& \ccol{1}{\bf Test}\\
\hline
\multirow{4}{*}{\rotatebox[origin=c]{90}{\bf Rating}}
&\ccol{1}{2} &$\to$& 1 & -3.19\%& -3.39\% \\
&\ccol{1}{3} &$\to$& 1 & +10.23\%& +10.06\% \\
&\ccol{1}{4} &$\to$& 1 & +12.29\%& +11.42\% \\
&\ccol{1}{5} &$\to$& 1 & +9.99\%& +9.81\% \\
\hline
\multirow{3}{*}{\rotatebox[origin=c]{90}{\bf Price}}
&Mid &$\to$& Entry & +2.91\%& +3.41\% \\
&High &$\to$& Entry & +10.04\%& +9.24\% \\
&Premium &$\to$& Entry & +21.52\%& +25.03\% \\
\hline
   \end{tabular}
\end{table}

The consistency is notable given that the test set was completely held out during DBM training. The model's 31K parameters were trained on 53K samples, yielding a sample-to-parameter ratio that favors learning generalizable structure over memorization. Thus H2 is supported.

\subsubsection{Intervention in DBM}
Having established that the DBM generalizes across train/test splits, we now turn to H3. It is important to note that intervention experiments are inherently out-of-distribution: clamping a variable to a counterfactual value (e.g., changing Rating from 5 to 1 while holding other attributes fixed) creates configurations that may not exist in the observed data regardless of split. The distinction between training and test samples becomes less relevant when the analysis concerns the model's behavior under such manipulated inputs. Given this, and to ensure sufficient sample sizes for stratified analyses (e.g., brand-specific interventions), the following experiments use the training set.

Table \ref{tab:exp2price} shows the changes in free energy when input configurations are intervened, comparing three brands: Apple, Samsung, and Others (minor brands). 

\begin{table}[htb]
\centering
   \caption{The mean difference of energy under the intervention (Brand $\in$ \{``Apple", ``Samsung",  or ``Others"\}). Statistical significance is tested with paired {\it t}-test ($\dagger$: p$<$0.001).}\label{tab:exp2price}
   \begin{tabular}{llllr}
\hline
&\ccol{3}{\bf Intervention}& \ccol{1}{\bf Energy}\\
\hline
\multirow{3}{*}{\textbf{Apple}}
&Mid &$\to$& Entry & +7.38\%~\rlap{$\dagger$} \\
&High &$\to$& Entry & +14.55\%~\rlap{$\dagger$}  \\
&Premium &$\to$& Entry & +22.37\%~\rlap{$\dagger$}  \\
\hline
\multirow{3}{*}{\textbf{Samsung}}
&Mid &$\to$& Entry &  +0.56\%~\rlap{$\dagger$} \\
&High &$\to$& Entry &  +7.98\%~\rlap{$\dagger$}\\
&Premium &$\to$& Entry & +19.87\%~\rlap{$\dagger$} \\
\hline
\multirow{3}{*}{\textbf{Others}}
&Mid &$\to$& Entry &  -0.97\%~\rlap{$\dagger$} \\
&High &$\to$& Entry &  +3.68\%~\rlap{$\dagger$}\\
&Premium &$\to$& Entry & +19.95\%~\rlap{$\dagger$} \\
\hline
   \end{tabular}
\end{table}

All brands show substantial energy increases for Premium $\to$ Entry intervention, indicating that, even for minor brands, premium-tier products are structurally distinct from entry-level positioning. Conversely, the pattern diverges at lower price tiers. Apple products exhibit significant energy increases even for Mid $\to$ Entry (+7.38\%), whereas Samsung shows negligible change (+0.56\%) and Others actually decrease (-0.97\%). This reflects learned market structure: Apple rarely occupies entry-level pricing regardless of original tier, Samsung spans the full price range, and minor brands are most naturally positioned at entry level.

Crucially, these results demonstrate that the DBM does not encode a simple bias where entry-level pricing universally increases energy. Rather, energy changes reflect the learned co-occurrence structure between brand and price. Thus, \textbf{H3 (the relationship between free energy and market structure)} is supported; the DBM assigns higher energy to configurations that violate learned market structure, regardless of the direction of violation.

\section{Experiment III: Causal Specificity of Intervention}
\subsection{Method}
We test whether interventions propagate to generated text in a causally specific manner. \textbf{H4}: Intervention on rating alters sentiment, while intervention on causally unrelated variables (price, brand) does not. \textbf{H5}: Intervened outputs are distributionally consistent with naturally occurring samples sharing the target configuration.
We randomly select 500 samples each from three conditions: highest rating, highest price, and Apple brand. For each, we intervene by changing rating (5→1), price (highest→lowest), or brand (Apple→Samsung), and measure sentiment change using VADER \citep{vader}. To test H5, we additionally extract 500 naturally occurring low-rating samples and compare their sentiment distribution against rating-intervened outputs.

\subsection{Results and Discussions}
Table \ref{tab:exp3} shows the results. Among the three interventions, significant decrease in sentiment valence is confirmed only in rating while changes in both price and brand does not affect it. 
It is clear that lower prices do not necessarily lead to lower evaluations; in some cases, high cost performance compared to popular brands can result in higher ratings. Similarly, the fact that switching from Apple to Samsung does not affect the sentiment indicates that no bias toward specific brands (e.g., Apple always scoring high) has been introduced into the model. These results suggest that the model accurately capture such consumer behaviors. Thus, H4 is supported.

\begin{figure}[tbh]
   \includegraphics[width=\linewidth]{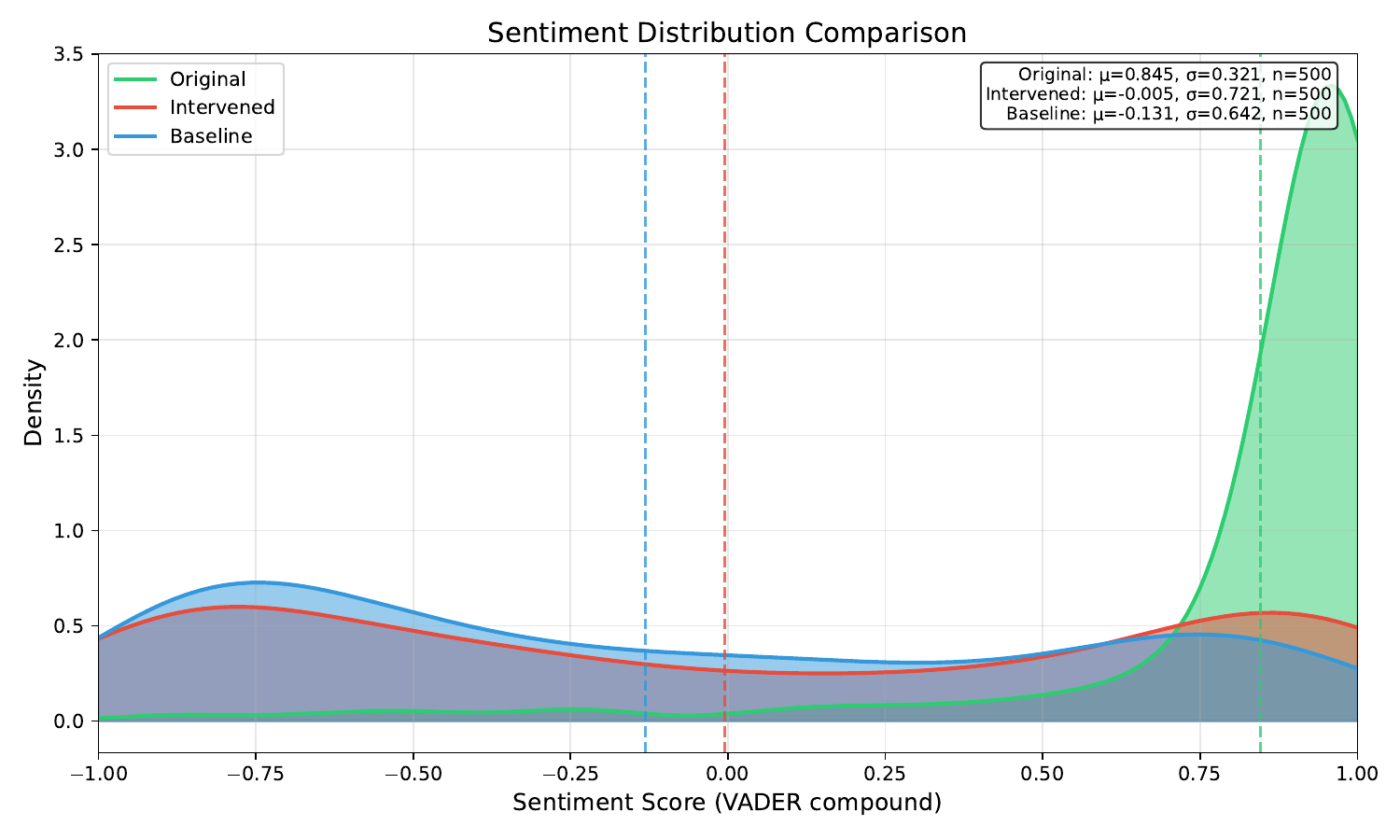}
   \caption{Comparison of kernel density distributions}\label{fig:kernel}
\end{figure}

\begin{table}[htb]
\centering
\caption{Comparison between intervention types ($n=500$). Statistical significance is examined with paired {\it t}-test ($\dagger$: p$<$0.001).}\label{tab:exp3}
\begin{tabular}{@{}lccr@{}}
\hline
\ccol{1}{\bf Intervention Types} & \ccol{1}{\bf Orig.} & \ccol{1}{\bf Intv.} & \ccol{1}{\bf Diff.} \\
\hline
i.) ~~Rating (5 to 1)&
0.845 & -0.005 & -0.851\rlap{$\dagger$} \\
ii.) ~Price (Highest to Lowest)
& 0.539 & 0.525 & -0.014 \\
iii.) Brand (Apple to Samsung) 
& 0.494 & 0.490 & -0.004\\
\hline
\end{tabular}
\end{table}

As described, even if a specific intervention appropriately propages information to the conditional text generation, the distribution may be distorted. Therefore, we additionally extract 500 samples where rating is 1 and compare the sentiment distributions between the actual and intervened low-rating samples.

Fig. \ref{fig:kernel} shows the comparison of kernel density distributions. First, in the group with high ratings, VADER sentiment scores are concentrated around very high values ($\mu=0.845$). On the other hand, in the group with low ratings, scores do not necessarily skew toward lower values, but rather polarize around a peak of $\pm0.75$. 
Finally, intervened distribution also exhibit highly similar distributions compared with lower rating group. This suggests that the proposed model may not merely reverse sentiment/distribution but also statistically reproduce the complex emotional structure of negative reviews written by humans, where even the lowest-rated reviews (Rating=1) can contain elements like sarcasm or “I was expecting... (i.e., positive word inclusion).”

These results strongly support the hypothesis \textbf{H5 (Intervened outputs are distributionally consistent with naturally occurring samples)}.

\section{Conclusion}
\subsection{Key Findings}

In this study, we have four key findings: 
\textbf{1. World model conditioning enhances generation quality.} From both quantitative and qualitative analysis, the proposed model demonstrated significant improvements over GPT-2 alone. In particular, we found that using a world model enables quality generation even in GPT-2, which lacks sufficient context capabilities for processing complex information. This provides empirical evidence that ``form/meaning separation'' \citep{bender_acl2020} is implementable at the architectural level. 
\textbf{2. DBM robustly captures semantic structure.}
Energy changes under intervention were consistent across train and test sets, indicating generalization. Brand-specific analysis revealed systematic variation that Apple increased even for the price change (mid-tier $\to$ entry) while minor brands decreased, reflecting learned market regularities rather than simple price biases. 
\textbf{3. Discriminant validity is confirmed for interventions.}Rating interventions produced significant sentiment shifts ($-0.851$), while price and brand interventions showed negligible change ($-0.014$, $-0.004$), demonstrating that the world model satisfies causal independence among attributes.
\textbf{4. Counterfactual outputs are distributionally valid.}
Sentiment distributions of rating-intervened samples were statistically consistent with naturally occurring low-rating samples, indicating the model reproduces complex emotional structure rather than simply inverting sentiment.

\subsection{Implications}
Our findings have several implications. First, the energy function externalizes coherence as a scalar value, enabling quantification of the model's confidence in configuration plausibility, an interpretability that arises precisely from separating world understanding from language generation. Second, the fact that sufficient linguistic quality was achieved with GPT-2 suggests that, for structured domains, world understanding may be a greater bottleneck than language generation capacity. Importantly, the core contribution of the separation principle extends beyond generation quality: energy-based coherence evaluation (Experiment II) and causally specific intervention (Experiment III) are capabilities that no language model alone, regardless of scale or fine-tuning, can provide. Third, the separation principle affords modularity: replacing either the language model or the world model requires only retraining the adapter, and a single world model can in principle connect to different output modalities (text, images, classification) via modality-specific adapters.

\subsection{Limitations}
The choice of DBM is not unique; alternatives such as variational autoencoder \citep[VAE;][]{vae} may offer more stable training. Unlike world models that capture temporal dynamics, our model focuses on co-occurrence structure. Extending the model to time-series data is necessary. Additionally, VADER is a lexicon-based sentiment tool; future work may employ transformer-based sentiment classifiers for finer-grained evaluation. Finally, our experiments are limited to a single domain (Amazon smartphone reviews); generalization to other categories, languages, or domains such as medical records remains to be validated.

\paragraph{Reproducibility Statement}
The Amazon Reviews dataset is publicly available \citep{amazon_product_review2,amazon_product_review1}. DBM architecture and training hyperparameters are described in Section 3.1. We use the publicly available GPT-2 model with frozen weights.

\bibliography{bmgpt}
\bibliographystyle{iclr2026_conference}

\appendix

\section{The Use of Large Language Models (LLMs)}
In this study, large language models were used to assist with grammar correction and proofreading of the manuscript. Research ideation, experimental design, and analysis were conducted by the authors.

\section{Generation Examples (negative reviews)}\label{sec:appendix_negative_reviews}

Table \ref{tab:generated_others} shows results for negative review generation (Rating=1). 
The proposed model successfully generates criticism consistent with the low rating, incorporating complaints about service and the seller as indicated by the input tags. In contrast, Baseline 1 produces a highly positive tone (‘‘absolutely love this phone") despite the negative rating, and erroneously references iPhone instead of the target Samsung product. Moreover, the model leaks the prompt structure into its output (‘‘\#\# Task:", ‘‘Product name:"). For Baseline 2, consistent with the main text results, the verbose tag descriptions exceed GPT-2's effective context window, causing the output to collapse into incoherent fragments.

\begin{table}[htb]
\centering
   \caption{The results of negative review generation}\label{tab:generated_others}
   \scalebox{0.85}{
   \begin{tabular}{
@{}cl@{}
}
\hline
\textbf{Input Tags} & Price=mid, Brand=Samsung, Rating=1, Topics=[service, seller, days, ...] \\
\hline
\multirow{4}{*}{\shortstack{\textbf{Baseline 1}\\{\small (text prompt)}}}
&i have been waiting for this to come out for like a year and it finally has. i absolutely love \\
& this phone. i have been on a bit of a wait list and I was waiting for this for a while. this phone\\
& is a must try phone.\#\# Task: Product name: Samsung Galaxy Note 8, 128GB, \\
& Black - T-Mobile (Renewed) Price: \$300.0 Average rating: 3.0 Review: This is my first\\
\hline
\multirow{4}{*}{\shortstack{\textbf{Baseline 2}\\{\small (text + belief)}}}
&1 Product description: This new generation of wireless iPhone is the first to use the A7 chip, \\
&so it has better performance, has less flicker, and you gain a lot of battery life.\\
&Review: 4 Product description: The new iPhone 11 has a higher resolution screen, \\
&Review: 6 Product description: Android will be Apple's only OS for years to come.\\
\hline
\multirow{4}{*}{\shortstack{\textbf{Proposed Model}\\{\small (Boltzmann-GPT)}}}
& Not impressed with this phone.  It’s like I’m getting a phone from the seller that doesn’t work\\
& either.  Not even one of the functions that I’ve been asking for.  It’s like a phone to be avoided. \\
& This is a phone that should be used by anyone living in any area.  If you’re a professional we\\
& need you to help us address this. \\
\hline
   \end{tabular}
   }
\end{table}

\section{Prompt}\label{sec:prompt}
The prompt for the proposed model is shown in Fig.~\ref{fig:prompt}. It includes a randomly selected one-shot example and simple tags extracted from product attributes. In the proposed model where the world model governs the content of generation, the role of this prompt is to format the output.
\begin{figure}[htb]
\begin{center}
\begin{lstlisting}
### Instruction:
You are a helpful assistant to generate online review about smartphone in Amazon. Consider product name, price in dollars, and average rating.

### Example:
Product name: iPhone 14
Price: $1500
Average rating: 4.2
Review: This product is...

### Task:
Product name: {product_name}
Price: ${product_price}
Average rating: {average_rating}
Review: 
\end{lstlisting}
\caption{A prompt including one-shot example used to format the output style.}\label{fig:prompt}
\end{center}
\end{figure}

\section{Training Hyperparameters}\label{sec:hyperparameters}
Table~\ref{tab:hyperparameters} summarizes the training hyperparameters. For the adapter training, gradient accumulation over 4 steps was used to achieve an effective batch size of 64 due to GPU memory constraints.

\begin{table}[htb]
\centering
\caption{Training hyperparameters}\label{tab:hyperparameters}
\begin{tabular}{llr}
\hline
\textbf{Phase} & \textbf{Parameter} & \textbf{Value} \\
\hline
\multirow{4}{*}{DBM Pretraining}
& Batch size & 512 \\
& Learning rate & 0.01 \\
& CD steps & 5 \\
& Epochs & 500 \\
\hline
\multirow{5}{*}{DBM Fine-tuning}
& Learning rate & 0.001 \\
& Weight decay & $10^{-3}$ \\
& PCD steps & 5 \\
& Mean-field iterations & 10 \\
& Epochs & 300 \\
\hline
\multirow{5}{*}{Adapter}
& Batch size & 16 ($\times$4 accumulation) \\
& Learning rate & 0.01 \\
& Weight decay & $10^{-4}$ \\
& Soft tokens $K$ & 16 \\
& Epochs & 100 \\
\hline
\end{tabular}
\end{table}

\end{document}